\documentclass[letterpaper]{article} 
\usepackage{aaai25}  
\usepackage{times}  
\usepackage{helvet}  
\usepackage{courier}  
\usepackage[hyphens]{url}  
\usepackage{graphicx} 
\usepackage{natbib}  
\usepackage{caption} 
\usepackage{algorithm}
\usepackage{algorithmic}

\usepackage{newfloat}
\usepackage{listings}
\DeclareCaptionStyle{ruled}{labelfont=normalfont,labelsep=colon,strut=off} 
\floatstyle{ruled}
\newfloat{listing}{tb}{lst}{}
\floatname{listing}{Listing}
\usepackage{tikz}
\usepackage{amsmath}
\usepackage{multirow}
\usepackage{makecell}
\usepackage{booktabs}    
\usepackage{colortbl}
\usepackage{comment}
\newcommand{\brio}{BRIDO }

\definecolor{codegreen}{rgb}{0,0.6,0}
\definecolor{codegray}{rgb}{0.5,0.5,0.5}
\definecolor{codepurple}{rgb}{0.58,0,0.82}
\definecolor{backcolour}{rgb}{0.95,0.95,0.92}

\lstdefinestyle{mystyle}{
    backgroundcolor=\color{backcolour},   
    commentstyle=\color{codegreen},
    keywordstyle=\color{magenta},
    numberstyle=\tiny\color{codegray},
    stringstyle=\color{codepurple},
    basicstyle=\ttfamily\footnotesize,
    breakatwhitespace=false,         
    breaklines=true,                 
    captionpos=b,                    
    keepspaces=true,                 
    numbers=left,                    
    numbersep=5pt,                  
    showspaces=false,                
    showstringspaces=false,
    showtabs=false,                  
    tabsize=2
}

\title{BRIDO: Bringing Democratic Order to Abstractive Summarization}
\author{
    Junhyun Lee\equalcontrib\thanks{Corresponding author.}, 
    Harshith Goka\equalcontrib, 
    Hyeonmok Ko
}
\affiliations{
    Samsung Research\\

    \{junhyun8.lee, h9399.goka, felix.ko\}@samsung.com
}

\usepackage{bibentry}

\begin{document}
\maketitle

\begin{abstract}
Hallucination refers to the inaccurate, irrelevant, and inconsistent text generated from large language models (LLMs).
While the LLMs have shown great promise in a variety of tasks, the issue of hallucination still remains a major challenge for many practical uses.
In this paper, we tackle the issue of hallucination in abstract text summarization by mitigating exposure bias. 
Existing models targeted for exposure bias mitigation, namely BRIO, aim for better summarization quality in the ROUGE score. 
We propose a model that uses a similar exposure bias mitigation strategy but with a goal that is aligned with less hallucination.
We conjecture that among a group of candidate outputs, ones with hallucinations will comprise the minority of the whole group. 
That is, candidates with less similarity with others will have a higher chance of containing hallucinated content.
Our method uses this aspect and utilizes contrastive learning, incentivizing candidates with high \emph{inter-candidate} ROUGE scores.   
We performed experiments on the XSum and CNN/DM summarization datasets, and our method showed {6.25\% and 3.82\%} improvement, respectively, on the consistency G-Eval score over BRIO.


\end{abstract}

\section{Introduction}

Large language models (LLMs) have achieved great success in many natural language processing tasks, such as machine translation~\cite{vaswani2017attention, arivazhagan2019massively}, question answering~\cite{karpukhin2020dense, zhu2021retrieving}, and text summarization~\cite{bart, zhang2020pegasus}.
However, despite their impressive performance, it has been shown that LLMs exhibit ``hallucination,'' in which they generate texts that are factually inaccurate, irrelevant, or inconsistent with the provided context.
Hallucination persists even in nowadays much larger LLMs~\cite{achiam2023gpt, anil2023palm, touvron2023llama} which are known, or believed, to have an order of tens or hundreds of billions of parameters.
Since such models can be potentially used in safety-critical applications, such as medical diagnosis or financial forecasting, hallucination poses a serious threat to the reliability and trustworthiness of these systems.

Among the variety of reasons researchers attribute to hallucination, Wang and Sennrich~\cite{wang2020exposure} first claimed that exposure bias~\cite{bengio2015scheduled, ranzato2016sequence} plays an important role. 
The mitigation of exposure bias is a subject that has been extensively investigated, and BRIO~\cite{liu2022brio} was one of them introduced in the scope of abstractive text summarization.
Though not specifically in the context of hallucination, it has achieved the state-of-the-art ROUGE~\cite{rouge} score at the time.
Following the logic that exposure bias is a cause of hallucination, one would expect that BRIO would have less hallucination than its base models, PEGASUS~\cite{zhang2020pegasus} and BART~\cite{bart}.

\begin{table}[t]
\centering
\begin{tabular}{cccc}
\Xhline{3\arrayrulewidth} \\[-1.8ex]
Metric& Model & XSum & CNN/DM \\
\Xhline{3\arrayrulewidth} \\[-1.8ex]
\multirow{3}{*}{\shortstack{QAFactEval\\(LERC)}} & base & \bf{2.047} & \bf{4.551} \\
 & BRIO & 1.927 & 4.070  \\
 & ref. & \it{1.713} & \it{3.460} \\ 
\hline \\[-1.8ex]
\multirow{3}{*}{\shortstack{QAFactEval\\(F1)}} & base & \bf{20.3} & \bf{80.6}  \\ 
 & BRIO & 19.0 & 65.9   \\
 & ref. & \it{15.1} & \it{48.6}  \\ 
\hline\\[-1.8ex]
\multirow{3}{*}{\shortstack{QAFactEval\\(EM)}} & base  & \bf{11.5} & \bf{66.2} \\
 & BRIO  & 10.5 & 50.7 \\
 & ref.  & \it{7.8} & \it{33.4}\\
\hline\\[-1.8ex]
\multirow{3}{*}{\shortstack{QAFactEval\\(IsAns)}} & base  & \bf{63.0} & \bf{97.1} \\
 & BRIO  & 60.8 & 92.4 \\
 & ref.  & \it{57.0} & \it{85.2}\\
\hline\\[-1.8ex]
\multirow{3}{*}{\shortstack{G-Eval\\(consistency)}} & base  &  \textit{3.62} & \it{4.42} \\
 & BRIO  &  \textbf{3.70} & \bf{4.45} \\
 & ref. & \textbf{3.70} & 4.43\\
\Xhline{3\arrayrulewidth}
\end{tabular}
\caption{QAFactEval and G-Eval consistency results on Xsum and CNN/DM datasets for BRIO outputs, the base model outputs, and the reference summary. The base model is PEGASUS for the XSum dataset and BART for the CNN/DM dataset. The highest score is in bold and the lowest is in italics.}
\label{tab:prelim}
\end{table}

As a preliminary experiment, we first check the aforementioned conjecture by comparing the hallucination metrics of BRIO and its base models on two text summarization datasets, XSum~\cite{xsum} and CNN/DM~\cite{cnndm}. 
(Details on the metrics and the datasets to follow in the Experimental Setup section.)
We show the results in Table~\ref{tab:prelim}, together with the score of the reference summary. 
Contrary to the expectation that BRIO would have mitigated hallucination, it scored less than the base models in most evaluations. 
We argue that this is due to the low performance of the reference summary, which scored worst among the three in most metrics.
Note that BRIO (the abbreviation of ``Bringing Order to Abstractive Summarization'') utilizes the ranking of candidate summaries calculated as the similarity with respect to the reference summary.

With this information, we conclude that BRIO is not well-suited for hallucination mitigation, although its exposure bias reduction scheme is still valid. 
Thus, we introduce an alternative to BRIO, whose goal is to handle hallucination better while retaining the ability to alleviate exposure bias. 
With the observation that the reference summary scores poorly in hallucination metrics, we avoid using the reference as the sole grader of the candidate summaries. 
Additionally, we hypothesize that when we have a diverse group of candidates the hallucinated content would consist of the minority among them. 
Then we can order the candidates by the similarity among them (this group may or may not include the reference summary), with a higher rank for candidates more similar to others collectively.
With this, we were able to achieve a {6.25\% and 3.82\%} increase in the LLM-based hallucination measure for the XSum and CNN/DM datasets, respectively.
This paper builds upon the sequence-to-sequence (seq2seq) model BRIO and compares it with other seq2seq baselines.
However, the method can be applied to decoder-based LLMs and even greatly benefit from the large group of contemporary LLMs.

The main contribution of this work is as follows:
\begin{itemize}
    \item We introduce BRIDO, an abstract summarization model that mitigates hallucination via decreasing exposure bias. BRIDO employs rank-based contrastive learning where the ranking of the candidates is determined by the similarity among the candidate group. 
    \item We test BRIDO with two summarization datasets XSum and CNN/DM, and find a significant increase in the LLM-based hallucination measure.
\end{itemize}

\section{Related Work}

Hallucination of LLMs is a very active area of research and there are many useful survey papers on the subject matter~\cite{ji2023survey, zhang2023siren, rawte2023survey, huang2023survey, tonmoy2024comprehensive}.
Among the various aspects of hallucination, we would like to concentrate on the hallucination caused by exposure bias and its mitigation strategy.

Exposure bias~\cite{bengio2015scheduled, ranzato2016sequence} is a major source of hallucination in the pre-training stage~\cite{wang2020exposure}. 
LLMs have different procedures in training and inference. 
While training, the model learns the probability distribution between words (tokens) by a teacher-forced maximum likelihood estimation (MLE).
That is, regardless of the predicted token, the next token used in the following step is determined from the ground truth example. 
In contrast, during the autoregressive inference, the actual predicted (generated) token is included in the context of the next step.
This discrepancy between the training and inference is referred to the exposure bias. 

\citet{wang2023progressive} address hallucination in neural machine translation in the context of exposure bias. 
They demonstrate that introducing intermediate supervision signals from the source-like to target-like structure alleviates exposure bias and thus reduces hallucination. 
Other studies devoted to mitigating exposure bias include methods such as reinforcement learning~\cite{chen2019reinforcement}, unlikelihood training~\cite{welleck2019neural}, and minimum Bayes risk~\cite{bertsch2023s}, although they do not directly address the issue of hallucination.

One particularly important study on exposure bias mitigation related to our work is BRIO~\cite{liu2022brio}.
BRIO is an abstractive summarization model, whose goal is to maximize the ROUGE score by reducing exposure bias through contrastive learning.
The method was successful in the sense that it had achieved the state-of-the-art ROUGE score at the time. 
Since our proposed method is closely related to BRIO, we will present its mechanism in detail in the following section. 

Many other hallucination reduction strategies exists, which do not involve exposure bias, in all stages of the LLMs including dataset preparation, training, and inference.
We refer the readers to the survey papers for more information. 

\section{Bringing the Democratic Order}

Here we introduce our proposed method \textbf{BRIDO}, which stands for ``Bringing Democratic Order to Abstractive Summarization.''
Unlike BRIO, where the ``order'' is dictated by the ROUGE score only with the reference, BRIDO orders the candidates by collectively comparing the similarities with all other candidates, thus introducing a democratic taste to the ordering algorithm.


\paragraph{Motivation}

Let us first revisit the scheme of BRIO.
BRIO uses a contrastive loss in addition to the usual MLE loss of abstractive summarization.
For contrastive learning, $N$ candidates were generated through diverse beam search~\cite{diverse-beam-search} and ranked.
The ranking is based on the ROUGE score with the reference summary, which measures the lexical similarity.
It is the additional contrastive learning that complements the MLE loss, that mitigates the exposure bias. 
The method achieves the state-of-the-art ROUGE score (with the reference). 
This is exactly the information the contrastive loss is providing to the model -- to value a high ROUGE score. 

We now consider this strategy in the sense of hallucination mitigation and figure out points of modifications. 
First, we questioned whether the reference summary is good enough so that aligning with the reference also reduces hallucination. 
A preliminary experiment (Table~\ref{tab:prelim}) has been done to answer this question. 
The result is -- the reference consistently scored the lowest, and PEGASUS/BART scored the highest in all QAFactEval metrics.
The intuition that BRIO would be placed in between PEGASUS/BART and the reference was correct, but counter-intuitively, the reference score was much worse than PEGASUS/BART.
For G-Eval, BRIO performed better than the base models, but only by a small margin.

\begin{figure*}
    \centering
    \includegraphics[width=.9\linewidth]{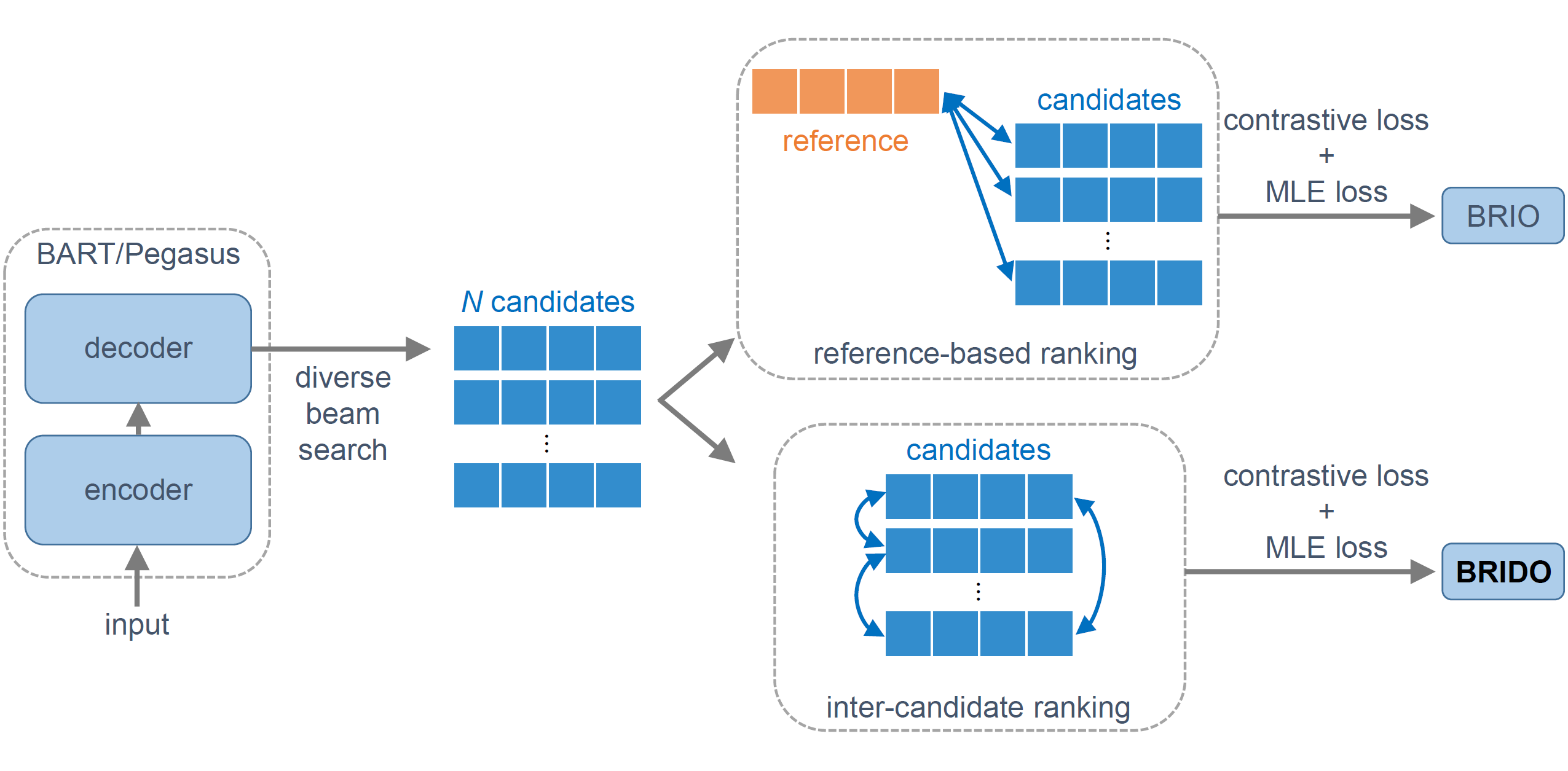}
    \caption{The BRIDO framework compared with BRIO~\cite{liu2022brio} and the base models (BART/Pegasus). BRIDO ranks the candidates based on the inter-candidate similarity, while BRIO uses the reference-based similarity score for ranking. In both cases, the similarity is measured by the ROUGE score.}
    \label{fig:brido}
\end{figure*}

This suggests that the reference is \textit{not} hallucination-free as assumed but is actually quite unfaithful. 
This is a characteristic of the XSum and CNN/DM datasets, as the reference there was collected from the bullet points of the news article. 
The bullet points are intended to complement the main article and are not an intended summary of the article.
Therefore they often include information that is not in the source, which is a hallucination when considered as a summary of the article.
This is also shown in previous works~\cite{liu2023revisiting} which compare the human evaluation of the reference and various model outputs. 
Consistent with our observation, the reference scored last in comparison with various models.

For example, in some op-ed articles the reference starts with ``[author]: ...'' which explicitly indicates this is the author's opinion and not the news outlet's official view. 
However, the author's name only shows up as a separate entity that is not part of the dataset and does not appear in the article (which is the source). 
In the perspective of the dataset, this is hallucination since it is an information that appeared in the summary which is not present in the source. 
Other examples include articles that are intended to be read after the title and the highlight. 
In those articles the title/highlight serves as a crucial context and contains information not included in the article (source). 
Instead, when the order is reversed where the article is the source and the highlight is its reference summary, then the reference will be inconsistant and inaccurate. 

To summarize, BRIO succeeded in the alignment with the reference and scored well in the ROUGE (which is a comparison with the reference), but unfortunately, the reference had a high level of hallucination.
Therefore it indeed reduced exposure bias, but not hallucination.
If we could find a better alternative to align with, we would be able to both reduce exposure bias and mitigate hallucination.



\subsection{Method}

Similar to BRIO, our method BRIDO is based on coordinating the model so that the model can assign a higher estimated probability to the ``better'' candidates during inference. 
This is not directly captured in the standard MLE objective used in training -- 
there could be multiple reasonable generations, and the MLE loss alone does not say anything about the order of the two imperfect references.
We require the model to be able to predict the ranking order of the set of most probable candidate summaries, which are its own beam search results. 

To achieve this objective, we train the model with a contrastive loss 
\cite{hopkins2011tuning, zhong2020extractive, liu2021simcls}: 
\begin{equation}
    \mathcal{L}_{\text {ctr}} = \sum_{i} \sum_{j>i} \max(0, f(S_j) - f(S_i) + \lambda_{ij}).
    \label{eq:crt_loss}
\end{equation}
$S_i$ and $S_j$ are candidate summaries that satisfy $Score(S_i) > Score(S_j)$, that is, $i$, $j$ are the rankings of the candidates measured in $Score$. 
$f(S_i)$ is the length-normalized estimated log probability, i.e., the estimated log probability normalized by $|S_i|^\beta$. ($\beta$ is the length penalty hyperparameter)
$\lambda_{ij}$ defines the target margin which is related to either the difference in score, $\lambda_{ij} = (Score(S_i) - Score(S_j)) * \lambda$, or difference in ranking, $\lambda_{ij} = (j - i)) * \lambda$, with the margin hyperparameter $\lambda$. 
%
%

Eq.~\eqref{eq:crt_loss} is common for both BRIO and BRIDO.
The main technical difference between the two is the scoring scheme, $Score$.
BRIO uses the ROUGE~\cite{rouge} score with the reference summary $S^*$ to score the candidate summaries. Defining $R(x,y)$ as the ROUGE score between $x$ and $y$, $Score_\textrm{BRIO}(S) = R(S, S^*)$. 
In BRIDO, we instead score these candidates by calculating the ROUGE score with the rest of the candidates and references, i.e., we use the \emph{inter-candidate} ROUGE: 
%
\begin{align}
\label{eq:alpha}
     Score&_\textrm{BRIDO}(S_i) = \nonumber \\
     &\frac{1}{N - 1 + \alpha} \left( \sum_{j\neq i} R(
S_i, S_j)  + \alpha R(S_i, S^*)\right).
\end{align}
%
$\alpha$ is a hyperparameter controlling the contribution between other candidate summaries and the reference. 
For the ROUGE calculation, we use the Python implementation (https://github.com/pltrdy/rouge) with the default settings. 

The intuition behind this choice of scoring scheme is that hallucinations would consist of a minority of information among a large enough group.
Consider a sentence with a single fact masked. 
If several models try to predict the fact, there would be one correct answer and a nearly infinite number of incorrect answers.
Therefore, if the fact can be inferred from the context and not from a random guess, the true fact would consist of the majority of the answers. 
In contrast, in the worst-case scenario of a random guesses, the correct answer and each of the different incorrect answers would occur in equal frequency. 
Thus using the scoring scheme of Eq.~\eqref{eq:alpha}, we conjecture that the candidates with high factuality will get high scores since they will agree more with other candidates.
Consequently, if we perform contrastive learning based on this ranking, we will be able to reduce both exposure bias and hallucination. 
Let us reiterate the conceptual similarities and distinctions between BRIO and BRIDO at this point.
They both use contrastive learning (as the multi-task together with the cross-entropy loss) to teach the models which summary is better, thus alleviating both exposure biases.
However, the ``better'' summary is the candidate with the higher $Score$ in each model. 
BRIO considers a higher $Score_\textrm{BRIO}$ as better, which implies a higher ROUGE score but not necessarily less hallucination. 
In contrast, a high $Score_\textrm{BRIDO}$ is considered better in BRIDO, and together with the aforementioned conjecture this directly implies less hallucination.

An alternative choice of scoring can be a direct measure of hallucination, for example using the hallucination metric used in this paper (QAFactEval and G-Eval) as the scoring scheme. 
This is similar in spirit to BRIO -- to maximize ROUGE, use ROUGE as the score -- and has the advantage that it does not need any additional assumptions. 
However, hallucination metrics require additional models which means much more computation and the possibility of error propagation.
Here, to limit the additional cost and to ensure the efficiency of the method, we choose to proceed with a simple scheme (ROUGE) together with a reasonable assumption. 
Note that ROUGE can be calculated between \emph{any} two texts, although it has been conventionally used as that between summary and reference. 
In the BRIDO scoring scheme, we are considering ROUGE between different summary candidates. 

The contrastive loss (Eq. \eqref{eq:crt_loss}) is added to the cross-entropy loss. 
The total loss becomes:
%
\begin{equation}
    \mathcal{L} =\mathcal{L}_{\text{xent}}+\gamma \mathcal{L}_{\text{ctr}}.
    \label{eq:loss}
\end{equation}
Here, $\gamma$ is the weight of the contrastive loss compared to that of the cross-entropy loss.
With this multi-task objective, BRIDO learns the ability to judge which candidate is ``better'' (less exposure bias) than just to summarize.
And, the ``better'' summarization for BRIDO is that with less hallucination.


Note that apart from the scoring scheme, our method is conceptually analogous to BRIO.
A schematic illustration of BRIDO, in comparison with BRIO and the base models, is shown in Figure \ref{fig:brido}.

\subsection{Hyperparameters}

We enumerate the hyperparameters in training BRIDO. 
We leave the inference parameters such as temperature, top-$k$, and top-$p$ unchanged from BRIO and do not list them here.
While generating the candidates from the pre-trained models, the diverse beam search~\cite{diverse-beam-search} is used. It has three main parameters -- the diversity penalty($\eta$), the number of beam groups($N_g$), and the number of candidates ($N$).
When defining the contrastive loss (Eq.~\eqref{eq:crt_loss}) and the score within, there are two more hyperparameters -- the margin value ($\lambda$) and the weight of reference summary ($\alpha$).

\paragraph{Diversity penalty ($\eta$)} 
The diversity penalty ensures that the outputs of the diverse beam search are different enough. 
The larger the value, the more diverse the outputs.
\paragraph{Number of beam groups ($N_g$)}
The total number of candidates is divided into $N_g$ beam groups. 
The diversity penalty applies to outputs between distinct beam groups. 
Candidates within the beam group are conventional beam search results. 
\paragraph{Number of candidates ($N$)} The total number of candidates to generate. We have two settings on the relation between $N$ and $N_g$: one is ``number of candidates / 8 $=$ number of beam groups'' i.e., $\frac{N}{8} = N_g$ used in XSum experiments, and the other is ``number of candidates $=$ number of beam groups.'' i.e., $N = N_g$ used in CNN/DM experiments.
That is, each beam group has 8 candidates in XSum, and a single candidate in CNN/DM. 
\paragraph{The role of reference summary ($\alpha$)} 
We can choose to include the reference summary in the scoring process. Without the reference, the model's outcome is very much dependent on the base model's performance and, therefore, may be unstable. We control the involvement of reference with a hyperparameter $\alpha$, which is defined in Eq. \eqref{eq:alpha}. 
%
$\alpha = 0$ is the reference-less limit, and $\alpha \rightarrow \infty$ is the BRIO limit.
In the experiments, we consider two more cases where $\alpha = 1$ (reference is just another candidate) and $\alpha = N - 1$ (the final score is the macro-average of the inter-candidate ROUGE and the reference ROUGE).

\begin{table*}[t]
    \centering
\begin{tabular}{lcccccccccccc}
\Xhline{3\arrayrulewidth} \\[-1.8ex]
\multirow{2}{*}{Dataset} &  \multirow{2}{*}{Model} & \multicolumn{3}{c}{ROUGE} & & \multicolumn{4}{c}{QAFactEval} & & \multicolumn{2}{c}{G-Eval} \\
\cline{3-5} \cline{7-10} \cline{12-13}\\[-1.8ex]
& & R-1 & R-2 & R-L & & LERC & F1 & EM & IsAns & &  cons. & ave. \\
\Xhline{3\arrayrulewidth} \\[-1.8ex]
\multirow{3}{*}{XSum} & {PEGASUS} &  47.18 & 24.62 & 39.37 & &\textbf{ 2.047} & 20.29 & 11.46 & 62.97 &&  3.62 & 3.31 \\ 
&{BRIO} & \textbf{48.86} & \textbf{25.54} & \textbf{40.48} & & 1.956 & 18.98 & 10.47 & 61.50  & &3.68 & 3.37  \\
& {BRIDO} & 47.69 & 24.62 & 39.19 & & 2.028 & \textbf{20.44} & \textbf{11.71} & \textbf{63.08} && \textbf{3.80} & \textbf{3.47} \\ 
\hline \\[-1.8ex]
\multirow{3}{*}{CNN/DM} & {BART} & 44.24 & 21.30 & 41.06 & & \textbf{4.551} & \textbf{80.56} & \textbf{66.17} & \textbf{97.08} && 4.42 & 3.82 \\
& {BRIO} & \textbf{47.74} & \textbf{23.75} & \textbf{44.55} & & 4.070 & 65.88 & 50.72 & 92.38 && 4.45 & 3.91 \\
& {BRIDO} & 45.81  & 22.95 & 42.51 & & 4.392 & 75.45 & 60.64  & 95.45 && \textbf{4.55} & \textbf{3.93} \\ 
\Xhline{3\arrayrulewidth}
    \end{tabular}
    \caption{The base results of BRIDO, in comparison with the baseline models PEGASUS/BART and BRIO, for the XSum and CNN/DM datasets. BRIDO improves over BRIO in all G-Eval and QAFactEval metrics. }
    \label{tab:brido}
\end{table*}

\paragraph{Margin value ($\lambda$)}
The margin value is defined in $\lambda_{ij}$ which is mentioned in Eq.~\eqref{eq:crt_loss}. We consider two types of margins: fixed margin (the type used in BRIO) and difference margin. 
Fixed margin between two candidates $S_i$ and $S_j$ is defined as $\lambda_{ij} = (j - i) * \lambda$, where the subscripts $i$ and $j$ indicates the ranking based on the score ($Score(S_i) > Score(S_j)$). 
Whereas the difference margin between these two candidates is defined as $\lambda_{ij} = (Score(S_i) - Score(S_j)) * \lambda$. 
The fixed margin drives the model to have the length-normalized estimated log probability at least equality separated based on their ranking. 
In the difference margin scheme the separation is proportional to the score of the candidates, and thus can be considered to  have higher resolution. 

\section{Experimental Setup}
\label{sec:setup}

The BRIDO model is fine-tuned with the XSum \cite{xsum} and CNN/DM \cite{cnndm} datasets to reduce the loss of Eq.~\eqref{eq:loss} from the corresponding pre-trained models, PEGASUS \cite{zhang2020pegasus} and BART \cite{bart}. 
The same hyper-parameters and optimizer (learning rate = $2\times 10^{-3}$, epoch = 2, and Adam optimizer~\cite{kingma2014adam}) are used for all datasets to validate the universality of our method except for the following differences -- we use a batch size of 16 for PEGASUS - XSum experiments and a batch size of 24 for BART - CNN/DM experiments. We keep the same generation parameters as BRIO during the evaluation for each dataset. 
All training and evaluation are conducted using 8 Nvidia Ampere A100 GPUs (80GB).

\subsection{Dataset}
We use the following two datasets widely used in abstractive text summarization in our experiments.

\paragraph{XSum} \cite{xsum} is a highly abstractive dataset of articles from the British Broadcasting Corporation (BBC), and the introductory sentences (generally one sentence) in these articles are regarded as summaries. 
\paragraph{CNN/DM} \cite{cnndm} is a news dataset from the Cable News Network (CNN) and the Daily Mail (DM). Following \cite{nallapati-etal-2016-abstractive}, we treat the news articles as the source and the associated highlights as the summaries. 

\paragraph{}However, as previously noted, considering the introductory sentence (XSum) or the article highlight (CNN/DM) as the article summary has hallucination issues.

\subsection{Metric}
We choose the metrics to cover the various summarization and hallucination metrics. 
\paragraph{ROUGE} Recall-Oriented Understudy for Gisting Evaluation \cite{rouge} measures the lexical similarity between summaries and human-written references by comparing the overlap of $n$-grams, which are sequences of $n$ words. The most common variants of ROUGE include ROUGE-N (for $n$-gram matching) and ROUGE-L (for longest common subsequence matching). 
Among these, we use ROUGE-1 (R-1), ROUGE-2 (R-2), and ROUGE-L (R-L) for evaluation.
For the ranking, we use $2 ~\textrm{R-1} \times \textrm{R-2} / (\textrm{R-1} + \textrm{R-2})$ for XSum and $(\textrm{R-1} + \textrm{R-2}+ \textrm{R-L})/3$ for CNN/DM following \citet{liu2022brio}.
ROUGE's limitation includes the lexical bias and the questionable quality of the reference.
In this work, we generalize this conventional definition of ROUGE and use it to measure the lexical similarity between summary candidates as well.

\begin{table*}[t]
    \centering
\begin{tabular}{lcccccccccccc}
\Xhline{3\arrayrulewidth} \\[-1.8ex]
\multirow{2}{*}{Dataset} &  \multirow{2}{*}{$\eta$} & \multicolumn{3}{c}{ROUGE} & & \multicolumn{4}{c}{QAFactEval} & & \multicolumn{2}{c}{G-Eval} \\
\cline{3-5} \cline{7-10} \cline{12-13}\\[-1.8ex]
& & R-1 & R-2 & R-L & & LERC & F1 & EM & IsAns &  & cons. & ave. \\
\Xhline{3\arrayrulewidth} \\[-1.8ex]
\multirow{3}{*}{XSum} & {0.1} & \textbf{47.69} & \textbf{24.62} & \textbf{39.19} & & \textbf{2.028} & 20.44 & 11.71 & \textbf{63.08} &  & 3.80 & 3.47 \\ 
& {0.3} & 47.33 & 24.11 & 38.57 & & 2.010 & 20.50 & \textbf{11.84} & 62.96 & & \textbf{3.89} & \textbf{3.56}  \\
& {1.0} & 46.94 & 23.77 & 38.12 & & 2.007 & \textbf{20.52} & 11.81 & 62.85 &  & 3.86 & 3.55 \\ 
\hline \\[-1.8ex]
\multirow{5}{*}{CNN/DM} & {0.1} & 45.98 & 22.97 & 42.51 & & 4.414 & 76.55 & \textbf{62.18} & 95.62 &  & 4.55 & 3.94 \\
& {0.3} & \textbf{46.06} & \textbf{23.07} & 42.66 & & 4.379 & 75.21 & 60.56 & 95.35 & & 4.51 & 3.92 \\
& {1.0} & 45.81  & 22.95 & 42.51 & & 4.392 & 75.45 & 60.64  & 95.45 &  & 4.55 & 3.93 \\
& {3.0} & 46.03 & 22.99 & \textbf{42.69} & & \textbf{4.417} & 76.13 & 61.41 & \textbf{95.81} & & \textbf{4.58} & \textbf{3.95} \\
& {10.0} & 45.42 & 22.62 & 42.05 & & 4.413 & \textbf{76.64} & 62.17 & 95.65 & & 4.54 & 3.94 \\ 
\Xhline{3\arrayrulewidth}
    \end{tabular}
    \caption{The performance of BRIDO when tuning the diversity penalty, $\eta$. $\eta = 0.1$ and $\eta = 1.0$ are the same parameters as in Table~\ref{tab:brido} for XSum and CNN/DM, respectively. Considering our main metric G-Eval the best parameters are $\eta = 0.3$ (XSum) and $\eta = 3.0$ (CNN/DM).}
    \label{tab:eta}
\end{table*}

\paragraph{QAFactEval} \cite{qafacteval} is a QA-Based Factual Consistency Evaluation without a reference. Given a summary and the source document, it uses a question generation model to generate questions for each of the named entities. It then uses the source document as a reference for the question-answering model to generate the answers for each question. It evaluates the answers with the true answer and measures the consistency of a summary with its source document. 
The submetrics (LERC, F1, EM, and IsAns) are the methods used in comparing the model answer with the true answer.
LERC has a maximum score of 5 and F1, EM, and IsAns are measured in percent.
QAFactEval, like all other QA-based factuality metrics, suffers from bias towards extractive summaries.

\paragraph{G-Eval} \cite{geval} is a framework for using large language models with chain-of-thoughts (CoT) and a form-filling paradigm to assess the quality of NLG outputs. 
G-Eval consists of four sub-metrics which are coherence, consistency, fluency, and relevance. 
Among these four, consistency is the measure of hallucination. 
We use G-Eval consistency as a hallucination metric, and the average G-Eval score as a metric for the overall quality of the summarization.
This will be our main metric for both quality and hallucination, as it is assisted by a large auxiliary model and does not suffer from the aforementioned deficiencies of ROUGE and QAFactEval.
We use a slightly modified method by merging the quality assessment's various aspects into one prompt. We use GPT-4 \cite{achiam2023gpt} as our LLM evaluator.


\subsection{Baseline Models}

Following \citet{liu2022brio}, we use {PEGASUS} \cite{zhang2020pegasus} and the BRIO's XSum model as our baseline for the XSum experiments. 
Similarly, we use BART \cite{bart} and the BRIO's CNN/DM model as our baseline for the CNN/DM dataset experiments. Both {PEGASUS} and {BART} are large pre-trained seq2seq LLMs standard in the literature. 
BRIO's XSum and CNN/DM models use PEGASUS and BART as their base models, respectively.
In the following, we indicate both models as BRIO for simplicity. 

\section{Results}

We trained two BRIDO models, each finetuned with XSum and CNN/DM.
First, we fix the diversity penalty ($\eta$) to its corresponding BRIO value ($\eta = 0.1$ for XSum and $\eta = 1.0$ for CNN/DM) and also exclude the reference summary from the scoring ($\alpha = 0$).
We perform an extensive search for the remaining hyperparameters: $N \in \{16,\, 32\}$, $\gamma \in [10, \, 100]$, $\lambda \in [0.001, \, 0.1]$, and fixed/difference margin scheme. 

The best models were $N=32$, $\gamma = 50$, $\lambda = 0.01$, difference margin for XSum; $N=32$, $\gamma = 20$, $\lambda = 0.1$, difference margin for CNN/DM.
Their performance, together with the baseline evaluations are shown in Table~\ref{tab:brido}.
The results of the other experiments for the hyperparameter search is presented in the appendix.

The ROUGE score is best in BRIO for both experiments, as expected. 
However, it scored worst in all QAFactEval metrics by a wide margin which reiterates the results in Table~\ref{tab:prelim}. 
For the QAFactEval, the base models (PEGASUS/BART) were on average the best, however in part due to its extractive-ness of the summaries. 
PEGASUS and BART outputs tend to be more extractive than BRIO or BRIDO, and this is a reason for their high score in QAFactEval measures which suffers from bias towards extractive summaries. 

The fact that PEGASUS is not as dominant as BART in QAFactEval also supports that the extractiveness is the reason for the base models' high QAFactEval score.
Note that XSum is intended to be more abstractive than CNN/DM (that is, the XSum reference summary is more abstract than that of CNN/DM). 
This is also evident from the exceptionally high values of QAFactEval scores for BART on CNN/DM.
Scores such as 4.551 LERC and 97.08 IsAns cannot be explained without extractiveness having an effect.
That said, although BRIDO was not the best among the three in QAFactEval, it showed significant improvement from BRIO and impressively scored better than PEGASUS in three sub-metrics in XSum. 

G-Eval is where BRIDO was truly dominant. 
It scored best in both consistency and average for XSum and CNN/DM.
The improvement of the consistency score is 4.97\% (PEGASUS) and 3.26\% (BRIO) for XSum; 2.94\% (BART) and 2.25\% (BRIO) for CNN/DM.
Note that the improvement is more prominent for the more abstractive dataset XSum.
We conclude here that our method BRIDO has mitigated hallucination (high G-Eval consistency score) without any degradation in summarization ability (high G-Eval average score). 

\paragraph{Effect of diversity penalty} 
One crucial hyperparameter in BRIDO is the diversity penalty ($\eta$). 
This is because $\eta$ has a more sensitive optimal value than that in BRIO. 
$\eta$ should not be too small that the candidate summary is not diverse enough. 
Then the probability of the generated candidate having a hallucinated content would be too small for the model to learn its pattern.
In contrast, if $\eta$ is too large, the candidate's summarization quality may not be good enough as the diversity beam search has traded diversity with summarization quality. 
The optimal $\eta$ should be large enough to ensure the candidate summaries are diverse enough so that hallucinated content occurs regularly, and at the same time small enough that all candidates are above a certain quality.

\begin{table*}[t]
    \centering
\begin{tabular}{lcccccccccccc}
\Xhline{3\arrayrulewidth} \\[-1.8ex]
\multirow{2}{*}{Dataset} &  \multirow{2}{*}{$\alpha$} & \multicolumn{3}{c}{ROUGE} & & \multicolumn{4}{c}{QAFactEval} &  & \multicolumn{2}{c}{G-Eval} \\
\cline{3-5} \cline{7-10} \cline{12-13}\\[-1.8ex]
& & R-1 & R-2 & R-L & & LERC & F1 & EM & IsAns &  & cons. & ave. \\
\Xhline{3\arrayrulewidth} \\[-1.8ex]
\multirow{4}{*}{XSum} & {0} & 47.33 & 24.11 & 38.57 & & 2.010 & \textbf{20.50} & \textbf{11.84} & 62.96 &  & 3.89 & \textbf{3.56} \\ 
& 1  & 47.34 & 24.12 & 38.58 & & 2.010 & 20.46 & 11.79 & \textbf{63.02} &  & 3.88 & 3.55  \\
& 31  & 47.44 & 24.25 & 38.78 & & 2.008 & 20.27 & 11.55 & 62.89 & & \textbf{3.91} & 3.54 \\
& $\infty$   & \textbf{47.53} & \textbf{24.49} & \textbf{39.05} & & \textbf{2.012} & 20.15 & 11.49 & 62.85 & & 3.77 & 3.46 \\ 
\hline \\[-1.8ex]
\multirow{4}{*}{CNN/DM} & {0} & 46.03 & 22.99 & 42.69 & & \textbf{4.417} & \textbf{76.13} & \textbf{61.41} & \textbf{95.81} && 4.58 & 3.95 \\
& {1} & 46.10 & 23.03 & 42.76 & & 4.411 & 75.81 & 61.07 & 95.60 &  & 4.53 & 3.93 \\
& {31} & 46.65 & 22.98 & 43.00 & & 4.224 & 70.61 & 55.44 & 93.71 && \textbf{4.62} & \textbf{4.02} \\
&  $\infty$ & \textbf{47.21 } & \textbf{23.74} & \textbf{43.67} & & 4.211 & 70.25 & 55.31 & 93.52 &  & 4.52 & 3.97 \\ 
\Xhline{3\arrayrulewidth}
    \end{tabular}
    \caption{The performance of BRIDO when tuning the weight on reference summary, $\alpha$. The baseline results ($\alpha = 0$ row) are the same parameters as in Table~\ref{tab:eta} for $\eta = 0.3$ (XSum) and $\eta = 3.0$ (CNN/DM) which we have concluded the best parameters in the sense of hallucination mitigation. Considering our main hallucination metric G-Eval consistency, the best parameters are $\alpha = 31$ for both datasets.}
    \label{tab:alpha}
\end{table*}

We tune $\eta$ to search for the optimal value in both datasets.
For XSum we increase $\eta$ up to 1.0, and for CNN/DM we search from a range of 0.1 to 10.0. 
The results are shown in Table~\ref{tab:eta}.

We find some fluctuation in the performance. 
However, the performance difference between models was not significant compared to Table~\ref{tab:brido}.
The default value of $\eta$ in Table~\ref{tab:brido} is the value used in BRIO, which is already an optimized value from \citet{liu2022brio}.
We expected that the optimal value would be different for BRIO and BRIDO, as the two has quite distinct scoring scheme. 
While this expectation was indeed true, the difference was not as prominent.
We do observe some improvement in the G-Eval metrics and we chose $\eta = 0.3$ and $\eta = 3.0$ as the optimal parameter for XSum and CNN/DM, respectively.

\paragraph{Effect of the reference summary}
Now we turn to optimizing the scoring scheme and tune the role and weight of the reference summary, which is dictated by the hyperparameter $\alpha$. 
Recall that up to this point, the score was pure inter-candidate ROUGE and the reference summary was not included ($\alpha = 0$). 

We tune $\alpha$ in three additional values: 1, $N-1$, and $\infty$.
$\alpha = 1$ is where the weight of the reference summary is identical to the candidate summaries -- each summary, including the reference summary, casts one vote.
$\alpha = N-1$ is the average of the score of the candidate summaries and that of the reference summary. 
That is, the average of the $N-1$ candidate summaries gives a score, and the reference summary gives a score. 
The final score becomes a simple average between the two (similar to the macro-average).
Finally, $\alpha = \infty$ is when the reference summary dominates the scoring. 
This is the same setting with BRIO. 
Table~\ref{tab:alpha} shows the result of this experiment. 
Note that while $\alpha = \infty$ is the BRIO limit, the results of this is not identical to the BRIO result in Table~\ref{tab:brido} because of the different hyperparameters.

The result shows some performance improvement when the reference summary plays a role. 
For G-Eval consistency $\alpha = 31$ was the best for both XSum and CNN/DM.
Notice that the $\alpha = \infty$ BRIO limit has the lowest G-Eval consistency score among the four models. 
While this is not the same BRIO model in Table~\ref{tab:brido}, it demonstrates that the reference summary poses a negative role in hallucination reduction. 

The other metrics showed some trends in this experiment. 
For example, ROUGE monotonically increased as $\alpha$ increased. 
This is a direct consequence of the fact that ROUGE here is the ROUGE with the reference. 
The $\alpha = \infty$ ROUGE is smaller than that of the BRIO, but this is because BRIO was optimized with the ROUGE score while BRIDO was not. 
For the QAFactEval, the trend is reversed and smaller $\alpha$ tends to have better values. 

Our final best model parameters are $N=32$, $\gamma = 50$, $\lambda = 0.01$, $\eta = 0.3$, $\alpha = 31$, difference margin for XSum; $N=32$, $\gamma = 20$, $\lambda = 0.1$, $\eta = 3.0$, $\alpha = 31$, difference margin for CNN/DM.
The G-Eval consistency scores for those models are 3.91 and 4.62, respectively. 
This is a 6.25\% and 3.82\% improvement over BRIO on XSum and CNN/DM respectively, and an 8.01\% and 4.52\% improvement over the base models.

\section{Conclusion}

We have introduced BRIDO, a method building upon BRIO and adding a democratic taste to its ordering. 
This retains the strength of BRIO, which mitigates exposure bias and teaches the model how to rank different imperfect models, andw align the model to value less hallucination. 
This goal is achieved by the scoring scheme which includes the inter-candidate ROUGE.
We have conjectured that hallucinated content would consist the minority in a large enough candidate summaries, and therefore less hallucination would have a positive correlation with high inter-candidate ROUGE. 
The fact that the main objective of the model is aimed at less hallucination, and also the model results in less exposure bias both contributes to hallucination mitigation. 
We have shown the model play out with extensive hyperparameter search experiments and found that the hallucination measure (namely, the G-Eval consistency) has improved significantly.

While the hallucination metrics we have included in the manuscript, especially G-Eval, does measure effectively measure the degree of hallucination, human evaluation is still an important measure.
The literature suggest good correlation between G-Eval and human evaluation~\cite{geval}, and therefore we believe that our analysis would be in line with human evaluations as well.
Regardless, directly showing the human evaluation results is worth the investigation, and we plan to study the human evaluation on BRIDO as a future work. 

The BRIDO method is only applied to seq2seq models in the manuscript but a similar algorithm may be applied to, and greatly benefit from, decoder models as well.
With the plethora of different decoder models, together with a combination of many different parameter sizes, the BRIDO algorithm may turn out very effective in large decoder models.
Assuming that we have enough decoder models with above threshold output quality, we do not need to invoke the diverse beam search but rather use the outputs of various models to be the candidate summaries. 
This will ensure both the diversity and quality of the candidates, and result in a more powerful BRIDO model.

\bibliography{custom}


\appendix


\section{Sample Prompt for G-Eval}

Below is a sample prompt for G-Eval used for the summarization task.

\begin{quote}
    
You will be given one summary written for a source text.\\

Your task is to rate the summary in four metrics.
\\

Please make sure you read and understand these instructions carefully. Please keep this document open while reviewing, and refer to it as needed.\\

Evaluation Criteria:\\

Coherence (1-5) - the collective quality of all sentences. We align this dimension with the DUC quality question of structure and coherence whereby ``the summary should be well-structured and well-organized. 
The summary should not just be a heap of related information, but should build from sentence to a coherent body of information about a topic."\\

Consistency (1-5) - the factual alignment between the summary and the summarized source. 
A factually consistent summary contains only statements that are entailed by the source document.
Annotators were also asked to penalize summaries that contained hallucinated facts. \\

Fluency (1-3): the quality of the summary in terms of grammar, spelling, punctuation, word choice, and sentence structure.\\

Relevance (1-5) - selection of important content from the source. The summary should include only important information from the source document. 
Annotators were instructed to penalize summaries which contained redundancies and excess information.\\

Evaluation Steps:\\

1. Read the source text carefully and identify the main topic, main facts and details.\\

2. Read the summary and compare it to the news article.\\

3. Assign a score based on the Evaluation Criteria.\\

Source Text:\\
\{\{Document\}\}\\

Summary:\\
\{\{Summary\}\}\\

Evaluation Form (JSON only):\\
\{``coherence": 0, ``consistency": 0, ``fluency": 0, ``relevance": 0\}

\end{quote}

\onecolumn

\section{Code for Reference}
\label{appx:code}

The code is built upon BRIO's code \cite{briocode}.

The following code is the \texttt{build\_diverse\_beam} function to be replaced in \texttt{preprocess.py} file to calculate various scores for the generated candidates.



\begin{lstlisting}[language=Python]
def build_diverse_beam(input):
    (
        src_line,
        tgt_line,
        cands,
        src_line_untok,
        tgt_line_untok,
        cands_untok,
        tgt_dir,
        dataset,
        ranking_metric,
    ) = input
    cands = [sent_tokenize(x) for x in cands]
    abstract = sent_tokenize(tgt_line)
    _abstract = "\n".join(abstract)
    article = sent_tokenize(src_line)

    cache_initialized = False
    cache = None

    if dataset == "xsum":
        def joint_metric(score):
            return (
                2
                * score["rouge1"].fmeasure
                * score["rouge2"].fmeasure
                / (score["rouge1"].fmeasure + score["rouge2"].fmeasure)
                if (score["rouge1"].fmeasure + score["rouge2"].fmeasure)
                else 0
            )

    else:
        def joint_metric(score):
            return (
                score["rouge1"].fmeasure
                + score["rouge2"].fmeasure
                + score["rougeLsum"].fmeasure
            ) / 3

    if ranking_metric == "":
        def compute_ranking_metric(hyp, *args, **kwargs):
            score = all_scorer.score(_abstract, "\n".join(hyp))
            return joint_metric(score)

    elif ranking_metric == "intra_rouge":
        def compute_ranking_metric(
            hyp, index: int, all_hyps: List[str], *args, **kwargs
        ):
            nonlocal cache, cache_initialized
            if not cache_initialized:
                cache = np.zeros((len(cands), len(cands)), dtype=np.float32) - 1
                for idx1, hyp1 in enumerate(all_hyps):
                    for idx2, hyp2 in enumerate(all_hyps):
                        if idx1 == idx2:
                            continue
                        score = all_scorer.score("\n".join(hyp2), "\n".join(hyp1))
                        cache[idx1, idx2] = joint_metric(score)
                cache_initialized = True
            return float(cache[index][np.arange(len(cands)) != index].mean())

    elif ranking_metric == "intra_rouge_plus_ref":
        def compute_ranking_metric(
            hyp, index: int, all_hyps: List[str], *args, **kwargs
        ):
            all_hyps = all_hyps + [abstract]
            nonlocal cache, cache_initialized
            if not cache_initialized:
                cache = np.zeros((len(all_hyps), len(all_hyps)), dtype=np.float32) - 1
                for idx1, hyp1 in enumerate(all_hyps):
                    for idx2, hyp2 in enumerate(all_hyps):
                        if idx1 == idx2:
                            continue
                        score = all_scorer.score("\n".join(hyp2), "\n".join(hyp1))
                        cache[idx1, idx2] = joint_metric(score)
                cache_initialized = True
            return float(cache[index][np.arange(len(all_hyps)) != index].mean())

    elif ranking_metric == "intra_rouge_plus_ref_weighted":
        def compute_ranking_metric(
            hyp, index: int, all_hyps: List[str], weights: List[float], *args, **kwargs
        ):
            all_hyps = all_hyps + [abstract]
            weights_sum = sum(weights)
            weights = np.array([w / weights_sum for w in weights])
            nonlocal cache, cache_initialized
            if not cache_initialized:
                cache = np.zeros((len(all_hyps), len(all_hyps)), dtype=np.float32) - 1
                for idx1, hyp1 in enumerate(all_hyps):
                    for idx2, hyp2 in enumerate(all_hyps):
                        if idx1 == idx2:
                            continue
                        score = all_scorer.score("\n".join(hyp2), "\n".join(hyp1))
                        cache[idx1, idx2] = joint_metric(score)
                cache_initialized = True
            return float(
                ((cache[index] * weights)[np.arange(len(all_hyps)) != index]).mean()
            )

    weights = [1.0 for c in cands] + [float(len(cands) - 1)]
    candidates = [
        (x, compute_ranking_metric(x, idx, cands, weights))
        for idx, x in enumerate(cands)
    ]
    cands_untok = [sent_tokenize(x) for x in cands_untok]
    abstract_untok = sent_tokenize(tgt_line_untok)
    article_untok = sent_tokenize(src_line_untok)
    candidates_untok = [
        (cands_untok[i], candidates[i][1]) for i in range(len(candidates))
    ]
    output = {
        "article": article,
        "abstract": abstract,
        "candidates": candidates,
        "article_untok": article_untok,
        "abstract_untok": abstract_untok,
        "candidates_untok": candidates_untok,
    }
    with open(tgt_dir, "w") as f:
        json.dump(output, f)
\end{lstlisting}

\vspace{0.5cm}
The following code shows the changes made for using score difference in contrastive loss function. This functions is from the file \texttt{model.py}.
\vspace{0.5cm}

\begin{lstlisting}[language=Python]
def RankingLoss(
    score,
    summary_score=None,
    margin=0,
    gold_margin=0,
    gold_weight=1,
    no_gold=False,
    no_cand=False,
    use_ranking_score_as_margin=False,
    ranking_scores=None,
):
    ones = torch.ones_like(score)
    loss_func = torch.nn.MarginRankingLoss(0.0)
    TotalLoss = loss_func(score, score, ones)
    # candidate loss
    n = score.size(1)
    if not no_cand:
        for i in range(1, n):
            pos_score = score[:, :-i]
            neg_score = score[:, i:]
            pos_score = pos_score.contiguous().view(-1)
            neg_score = neg_score.contiguous().view(-1)
            ones = torch.ones_like(pos_score)
            if use_ranking_score_as_margin:
                pos_ranking_scores = ranking_scores[:, :-i]
                neg_ranking_scores = ranking_scores[:, i:]
                pos_ranking_scores = pos_ranking_scores.contiguous().view(-1)
                neg_ranking_scores = neg_ranking_scores.contiguous().view(-1)
                loss = torch.clamp(
                    -(ones * (pos_score - neg_score))
                    + (pos_ranking_scores - neg_ranking_scores) * margin,
                    0.0,
                ).mean()
            else:
                loss_func = torch.nn.MarginRankingLoss(margin * i)
                loss = loss_func(pos_score, neg_score, ones)
            TotalLoss += loss
    if no_gold:
        return TotalLoss
    # gold summary loss
    pos_score = summary_score.unsqueeze(-1).expand_as(score)
    neg_score = score
    pos_score = pos_score.contiguous().view(-1)
    neg_score = neg_score.contiguous().view(-1)
    ones = torch.ones_like(pos_score)
    loss_func = torch.nn.MarginRankingLoss(gold_margin)
    TotalLoss += gold_weight * loss_func(pos_score, neg_score, ones)
    return TotalLoss
\end{lstlisting}

\end{document}